\documentclass[letterpaper, 10 pt, conference]{ieeeconf}  

\IEEEoverridecommandlockouts                              

\usepackage{epsfig}
\usepackage{graphicx}
\usepackage{amsmath}
\usepackage{amssymb}
\usepackage{bbm}
\usepackage{booktabs}
\usepackage{multirow}
\usepackage{makecell}
\usepackage{xcolor}
\usepackage{xspace}
\usepackage{colortbl}
\usepackage[colorlinks=true,citecolor=green]{hyperref}

\title{\LARGE \bf
TeX-NeRF: Neural Radiance Fields from Pseudo-TeX Vision
}

\author{Chonghao Zhong$^{1}$, Chao Xu$^{1,\star}$ 
\thanks{$^{1}$ Chonghao Zhong and Chao Xu are with MoE Key Laboratory of Photo-electronic Imaging Technology and System, School of Optics and Photonics, Beijing Institute of Technology, Beijing, China.
{\tt\small zchnanguan7@gmail.com, rockyxu@bit.edu.cn}}%
\thanks{$^\star$ Chao Xu is the corresponding author.}
}

\makeatletter
\DeclareRobustCommand\onedot{\futurelet\@let@token\@onedot}
\def\@onedot{\ifx\@let@token.\else.\null\fi\xspace}

\makeatother

\begin{document}










\def\support{\mbox{support}}
\def\diag{\mbox{diag}}
\def\rank{\mbox{rank}}
\def\grad{\mbox{\text{grad}}}
\def\dist{\mbox{dist}}
\def\sgn{\mbox{sgn}}
\def\tr{\mbox{tr}}
\def\card{{\mbox{Card}}}

\def\balpha{\mbox{{\boldmath $\alpha$}}}
\def\bbeta{\mbox{{\boldmath $\beta$}}}
\def\bzeta{\mbox{{\boldmath $\zeta$}}}
\def\bgamma{\mbox{{\boldmath $\gamma$}}}
\def\bdelta{\mbox{{\boldmath $\delta$}}}
\def\bmu{\mbox{{\boldmath $\mu$}}}
\def\bftau{\mbox{{\boldmath $\tau$}}}
\def\beps{\mbox{{\boldmath $\epsilon$}}}
\def\blambda{\mbox{{\boldmath $\lambda$}}}
\def\bLambda{\mbox{{\boldmath $\Lambda$}}}
\def\bnu{\mbox{{\boldmath $\nu$}}}
\def\bomega{\mbox{{\boldmath $\omega$}}}
\def\bfeta{\mbox{{\boldmath $\eta$}}}
\def\bsigma{\mbox{{\boldmath $\sigma$}}}
\def\bzeta{\mbox{{\boldmath $\zeta$}}}
\def\bphi{\mbox{{\boldmath $\phi$}}}
\def\bxi{\mbox{{\boldmath $\xi$}}}
\def\bvphi{\mbox{{\boldmath $\phi$}}}
\def\bdelta{\mbox{{\boldmath $\delta$}}}
\def\bvarpi{\mbox{{\boldmath $\varpi$}}}
\def\bvarsigma{\mbox{{\boldmath $\varsigma$}}}
\def\bXi{\mbox{{\boldmath $\Xi$}}}
\def\bmW{\mbox{{\boldmath $\mW$}}}
\def\bmY{\mbox{{\boldmath $\mY$}}}

\def\bPi{\mbox{{\boldmath $\Pi$}}}

\def\bOmega{\mbox{{\boldmath $\Omega$}}}
\def\bDelta{\mbox{{\boldmath $\Delta$}}}
\def\bPi{\mbox{{\boldmath $\Pi$}}}
\def\bPsi{\mbox{{\boldmath $\Psi$}}}
\def\bSigma{\mbox{{\boldmath $\Sigma$}}}
\def\bUpsilon{\mbox{{\boldmath $\Upsilon$}}}

\def\mA{{\mathcal A}}
\def\mB{{\mathcal B}}
\def\mC{{\mathcal C}}
\def\mD{{\mathcal D}}
\def\mE{{\mathcal E}}
\def\mF{{\mathcal F}}
\def\mG{{\mathcal G}}
\def\mH{{\mathcal H}}
\def\mI{{\mathcal I}}
\def\mJ{{\mathcal J}}
\def\mK{{\mathcal K}}
\def\mL{{\mathcal L}}
\def\mM{{\mathcal M}}
\def\mN{{\mathcal N}}
\def\mO{{\mathcal O}}
\def\mP{{\mathcal P}}
\def\mQ{{\mathcal Q}}
\def\mR{{\mathcal R}}
\def\mS{{\mathcal S}}
\def\mT{{\mathcal T}}
\def\mU{{\mathcal U}}
\def\mV{{\mathcal V}}
\def\mW{{\mathcal W}}
\def\mX{{\mathcal X}}
\def\mY{{\mathcal Y}}
\def\mZ{{\mathcal{Z}}}


\def\bmA{{\mathbfcal A}}
\def\bmB{{\mathbfcal B}}
\def\bmC{{\mathbfcal C}}
\def\bmD{{\mathbfcal D}}
\def\bmE{{\mathbfcal E}}
\def\bmF{{\mathbfcal F}}
\def\bmG{{\mathbfcal G}}
\def\bmH{{\mathbfcal H}}
\def\bmI{{\mathbfcal I}}
\def\bmJ{{\mathbfcal J}}
\def\bmK{{\mathbfcal K}}
\def\bmL{{\mathbfcal L}}
\def\bmM{{\mathbfcal M}}
\def\bmN{{\mathbfcal N}}
\def\bmO{{\mathbfcal O}}
\def\bmP{{\mathbfcal P}}
\def\bmQ{{\mathbfcal Q}}
\def\bmR{{\mathbfcal R}}
\def\bmS{{\mathbfcal S}}
\def\bmT{{\mathbfcal T}}
\def\bmU{{\mathbfcal U}}
\def\bmV{{\mathbfcal V}}
\def\bmW{{\mathbfcal W}}
\def\bmX{{\mathbfcal X}}
\def\bmY{{\mathbfcal Y}}
\def\bmZ{{\mathbfcal Z}}

\def\0{{\bf 0}}
\def\1{{\bf 1}}

\def\bA{\boldsymbol{A}}
\def\bB{\boldsymbol{B}}
\def\bC{\boldsymbol{C}}
\def\bD{\boldsymbol{D}}
\def\bE{\boldsymbol{E}}
\def\bF{\boldsymbol{F}}
\def\bG{\boldsymbol{G}}
\def\bH{\boldsymbol{H}}
\def\bI{\boldsymbol{I}}
\def\bJ{\boldsymbol{J}}
\def\bK{\boldsymbol{K}}
\def\bL{\boldsymbol{L}}
\def\bM{\boldsymbol{M}}
\def\bN{\boldsymbol{N}}
\def\bO{\boldsymbol{O}}
\def\bP{\boldsymbol{P}}
\def\bQ{\boldsymbol{Q}}
\def\bR{\boldsymbol{R}}
\def\bS{\boldsymbol{S}}
\def\bT{\boldsymbol{T}}
\def\bU{\boldsymbol{U}}
\def\bV{\boldsymbol{V}}
\def\bW{\boldsymbol{W}}
\def\bX{\boldsymbol{X}}
\def\bY{\boldsymbol{Y}}
\def\bZ{\boldsymbol{Z}}

\def\ba{\boldsymbol{a}}
\def\bb{\boldsymbol{b}}
\def\bc{\boldsymbol{c}}
\def\bd{\boldsymbol{d}}
\def\be{\boldsymbol{e}}
\def\bff{\boldsymbol{f}}
\def\bg{\boldsymbol{g}}
\def\bh{\boldsymbol{h}}
\def\bi{\boldsymbol{i}}
\def\bj{\boldsymbol{j}}
\def\bk{\boldsymbol{k}}
\def\bl{\boldsymbol{l}}
\def\bn{\boldsymbol{n}}
\def\bo{\boldsymbol{o}}
\def\bp{\boldsymbol{p}}
\def\bq{\boldsymbol{q}}
\def\br{\boldsymbol{r}}
\def\bs{\boldsymbol{s}}
\def\bt{\boldsymbol{t}}
\def\bu{\boldsymbol{u}}
\def\bv{\boldsymbol{v}}
\def\bw{\boldsymbol{w}}
\def\bx{\boldsymbol{x}}
\def\by{\boldsymbol{y}}
\def\bz{\boldsymbol{z}}

\def\hy{\hat{y}}
\def\hby{\hat{{\bf y}}}

\def\mmE{{\mathbb E}}
\def\mmP{{\mathrm P}}
\def\mmB{{\mathrm B}}
\def\mmR{{\mathbb R}}
\def\mmV{{\mathbb V}}
\def\mmN{{\mathbb N}}
\def\mmZ{{\mathbb Z}}
\def\mMLr{{\mM_{\leq k}}}

\def\tC{\tilde{C}}
\def\tk{\tilde{r}}
\def\tJ{\tilde{J}}
\def\tbx{\tilde{\bx}}
\def\tbK{\tilde{\bK}}
\def\tL{\tilde{L}}
\def\tbPi{\mbox{{\boldmath $\tilde{\Pi}$}}}
\def\tw{{\bf \tilde{w}}}

\def\barx{\bar{\bx}}

\def\pd{{\succ\0}}
\def\psd{{\succeq\0}}
\def\vphi{\varphi}
\def\trsp{{\sf T}}

\def\mRMD{{\mathrm{D}}}
\def \DKL{{D_{KL}}}
\def\st{{\mathrm{s.t.}}}
\def\nth{{\mathrm{th}}}

\def\st{{\mathrm{s.t.}}}
\def\tr{\mathrm{tr}}
\def\grad{{\mathrm{grad}}}

\newtheorem{coll}{Corollary}
\newtheorem{deftn}{Definition}
\newtheorem{thm}{Theorem}
\newtheorem{prop}{Proposition}
\newtheorem{lemma}{Lemma}
\newtheorem{remark}{Remark}
\newtheorem{ass}{Assumption}

\def\red{\textcolor{red}}
\def\blue{\textcolor{blue}}




\maketitle
\thispagestyle{plain}
\pagestyle{plain}

\begin{abstract}

Neural radiance fields (NeRF) has gained significant attention for its exceptional visual effects. However, most existing NeRF methods reconstruct 3D scenes from RGB images captured by visible light cameras. In practical scenarios like darkness, low light, or bad weather, visible light cameras become ineffective. Therefore, we propose TeX-NeRF, a 3D reconstruction method using only infrared images, which introduces the object material emissivity as a priori, preprocesses the infrared images using Pseudo-TeX vision, and maps the temperatures (T), emissivities (e), and textures (X) of the scene into the saturation (S), hue (H), and value (V) channels of the HSV color space, respectively. Novel view synthesis using the processed images has yielded excellent results. Additionally, we introduce 3D-TeX Datasets, the first dataset comprising infrared images and their corresponding Pseudo-TeX vision images. Experiments demonstrate that our method not only matches the quality of scene reconstruction achieved with high-quality RGB images but also provides accurate temperature estimations for objects in the scene.

\end{abstract}

\section{INTRODUCTION}

Since its proposal, Neural Radiance Fields(NeRF)~\cite{nerf}  has achieved significant success in 3D reconstruction, novel view synthesis, and applications such as robotic perception and navigation~\cite{niceslam}\cite{latitude}\cite{vision}, virtual reality, and computer vision tasks like semantic segmentation~\cite{semanticray}~\cite{inplace}, target detection~\cite{nerfrpn}~\cite{mononerd}. Most NeRF models derive implicit scene representations from RGB images captured by visible cameras. Recently, NeRF has been extended to other modalities, including near-infrared, multispectral images~\cite{cross}, LiDAR point clouds~\cite{nerflidar} and even audio signals~\cite{novel}. Combining RGB with other modalities yields more accurate scene representations than using a single modality. However, thermal imagery is often used as a supplementary approach to enhance RGB images in low-light conditions. Multimodal NeRFs typically depend on significant or similar features between modalities, but infrared thermal imaging presents challenges such as low contrast, missing details, and blurring due to sensor noise, pixel array size, and the wavelength difference between infrared and optical radiation. These challenges complicate the effectiveness of Structure from Motion (SfM)-based camera pose reconstruction methods, impacting the quality of 3D reconstructions. While multimodal fusion, such as combining visible and infrared images or other sensor data, addresses some of these issues, it also increases system complexity, introduces data synchronization and calibration challenges, and raises computational overhead and power consumption.

\begin{figure}[!t]
    \centering
    \includegraphics[width=1\linewidth]{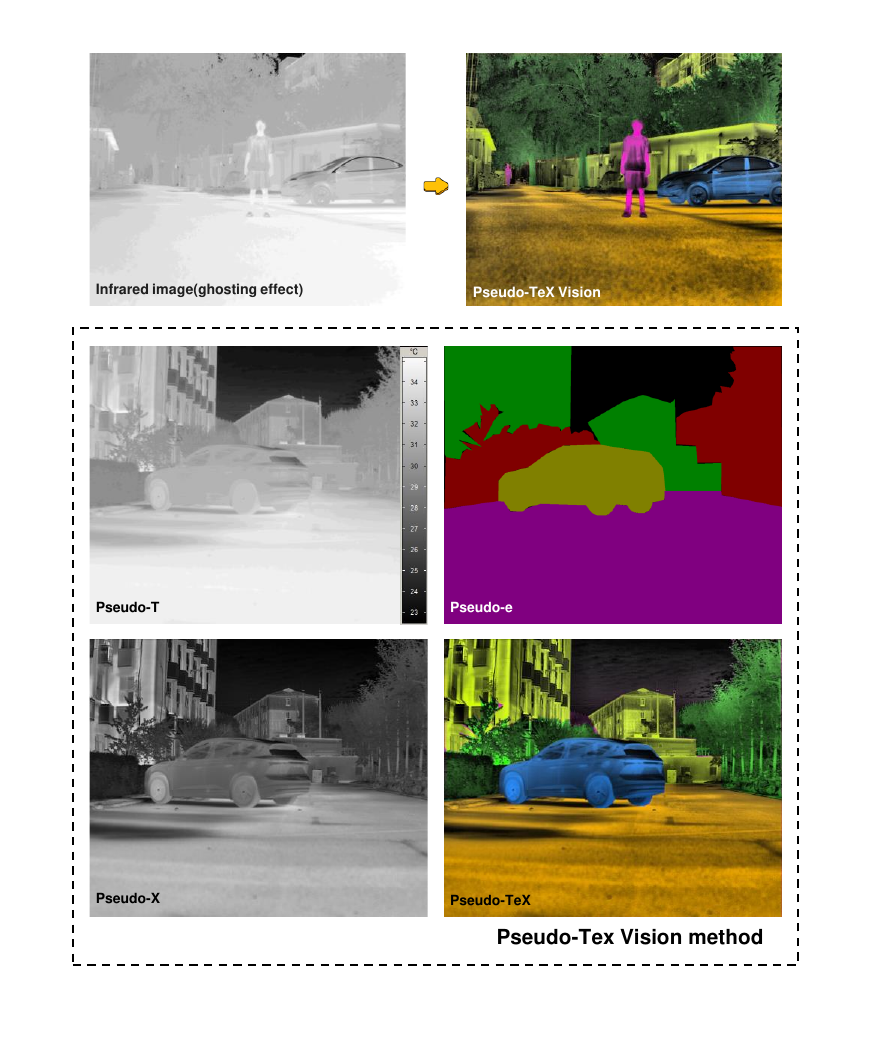}
    \vspace{-10mm}
    \caption{Pseudo-TeX Vision schematic. Enables infrared images to overcome ghosting effects and recover texture detail information.}
    \label{fig1}
    \vspace{-6mm}
\end{figure}

In this paper, we propose a high-quality 3D reconstruction method based on infrared images that requires data captured by only one type of sensor, an infrared thermal camera, combined with the average emissivity of the scene object in the infrared band as a priori, to obtain high-quality novel view synthetic quality and greatly improve the accuracy of the novel view's temperature estimation. We note that the total information in the thermal signal equation captures three physical quantities, temperature (T), emissivity (e), and texture (X). TeX vision uses hyperspectral heat cubes to decompose these quantities from the thermal signal and displays them in the HSV color space~\cite{hadar}. In RGB color space, each channel represents light intensity. In contrast, each channel in the HSV color space used in this paper has a physical meaning. The hue channel correlates with common object semantics (e.g. blue for water or sky, green for grass or leaves). In TeX vision, the saturation channel represents the image's temperature distribution (T), the hue channel indicates the material class (e), as material class is closely related to emissivity, and the value channel displays texture (X).

\begin{figure*}[!t]
  \centering
  \includegraphics[width=.85\linewidth]{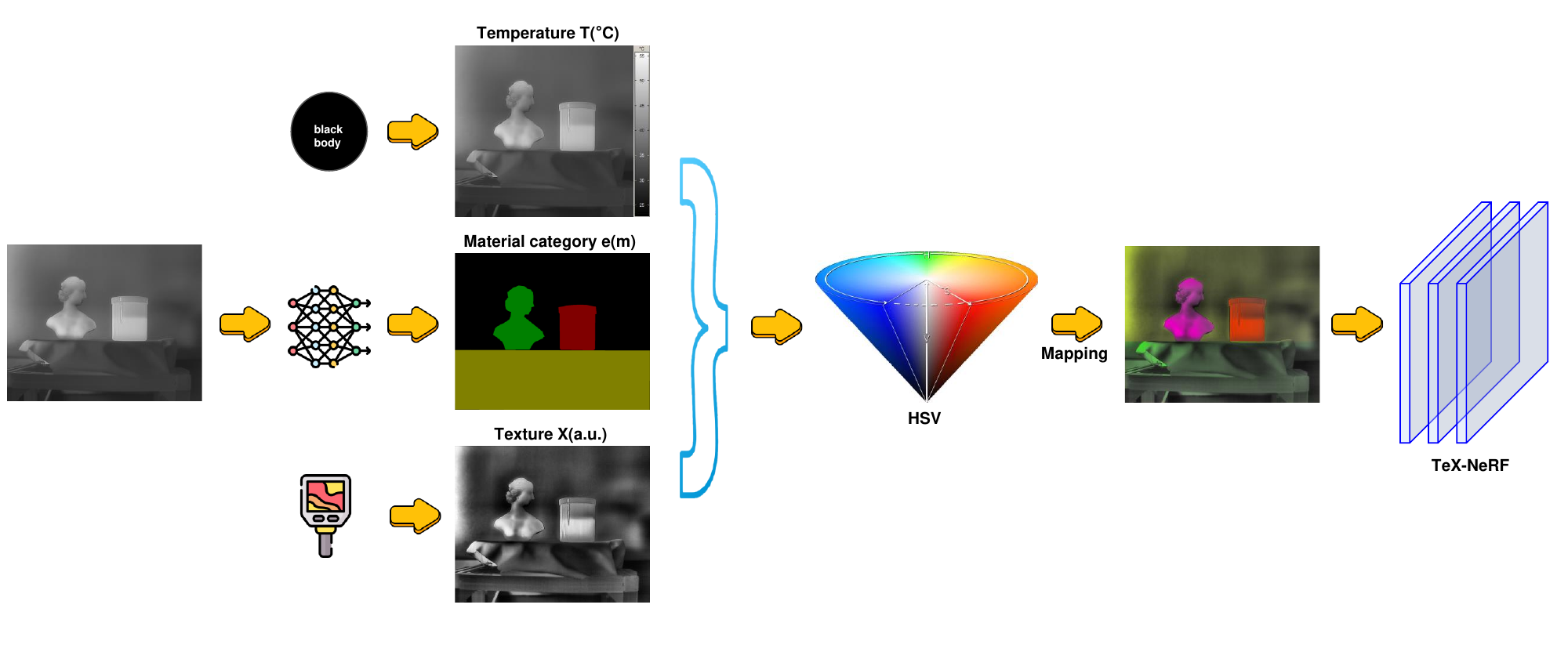}
  \vspace{-5mm}
  \caption{Overview of the proposed approach.
  The infrared image undergoes blackbody correction to extract temperature distribution information. Material data for each object in the scene is obtained through thermal vision-based semantic segmentation, combined with emissivity a priori information. Texture details are extracted using enhancement algorithms like the sensor's AGC. This data is mapped to the HSV color space to generate Pseudo-TeX Vision, and the processed image sequences are used as inputs for training TeX-NeRF.}
  \label{fig2}
  \vspace{-4mm}
\end{figure*}

In this study, our approach extends the application to common thermal datasets without spectral resolution. We use infrared thermal imaging to estimate temperature (T), adopt thermal vision-based semantic segmentation to approximate material categories e(m) by extracting spatial patterns and estimating semantic categories from thermal images, and apply infrared image detail enhancement and automatic gain control (AGC) to improve texture and visual contrast. These elements are combined to produce pseudo-TeX vision, as shown in the Fig.\ref{fig1}. Pseudo-TeX vision integrates spatial patterns, temperature differences, and detail enhancement to infer material and geometric information, making infrared images as rich in texture as visible light images with temperature information. In addition, we enhance NeRF by using HSV channel-based rendering instead of RGB channel-based rendering combined with pseudo-TeX vision-processed infrared images. An overview of the proposed approach is shown in Fig.\ref{fig2}. This approach can achieve high-quality scene reconstruction based on infrared images alone and achieves state-of-the-art accuracy in temperature estimation of synthesized novel views.

\section{RELATED WORK}

\subsection{Neural Radiance Fields (NeRF)}

NeRF generates high-fidelity 3D scenes and views by encoding the scene's rays using a multi-layer perceptron (MLP). The core idea of the technology is to learn the radiance field in 3D space from image data, enabling the synthesis of novel views from arbitrary viewpoints. NeRF uses MLPs to predict color and density based on the input 3D position and view direction, and integrates along camera rays to generate the final image. This approach enables NeRF to accurately capture complex lighting effects and details.

NeRF shows great potential in the synthesis and transformation of cross-modal data. In recent studies, NeRF has been extended to combine data from other modalities (e.g. depth maps, infrared images, etc.) with RGB images to generate richer 3D representations ~\cite{exploring}.ThermoNeRF~\cite{thermonerf} uses paired RGB and thermal images to learn the scene density, while using different networks to estimate color and temperature information to overcome the lack of texture in thermal images.Thermal NeRF~\cite{thermalnerf} proposes heat mapping for modeling thermal temperature values in the infrared and structural thermal constraints acting on thermal distributions, leveraging infrared features for high-fidelity 3D representation. Cross-modal extensions allow the system to create correlations between data with different sensors, thus enhancing the understanding and reproduction of 3D scenes.

It is worth mentioning that the proposal of 3D Gaussian Splatting~\cite{3dgs} further enhances the representation of NeRF in complex scenes. 3D Gaussian Splatting makes the representation of a scene more compact and easy to handle by representing each point in 3D space as a Gaussian distribution with position, color, and material properties. This approach not only reduces the complexity of data storage and computation, but also improves the quality and speed of rendering, enabling 3D scene representation to generate fine-grained scene views at higher resolutions.

\subsection{TeX Vision}

TeX Vision overcomes the ghosting effect, improves the accuracy of temperature estimation, improves object detection and ranging capabilities and provides additional physical attributes~\cite{why}. TeX Vision was first proposed in HADAR~\cite{hadar} and achieved stunning results. Emissivity is an important parameter of the object's radiation characteristics. TeX vision uses the object's radiation characteristics for perception, and obtains the emissivity information of different objects by constructing a library of emissivity materials in order to realize the identification of the object's material and the extraction of reflective features. In addition, TeX vision uses machine learning algorithms to process and analyze the collected data, and is able to realize the target's matching and identification. Further extract the texture and depth information of the target.

\subsection{Semantic Segmentation for Thermal Images}

In TeX Vision, hyperspectral thermal cubes are used as input, which in turn are computed to obtain the emissivity of the object, while in Pseudo-TeX Vision, we use semantic segmentation of thermal imaging images to obtain the material information of the scene, which is combined with the introduced material emissivity prior to obtain the average emissivity of the scene objects. In the field of semantic segmentation of thermal imaging images, researchers have developed various approaches to overcome the challenges due to thermal imaging characteristics such as edge blurring and sensor noise~\cite{semanticsurvey}.

EC-CNN~\cite{eccnn} leverages edge prior information to enhance segmentation quality. The model integrates an edge extractor, EC-CNN blocks, and a DeepLabV3-based~\cite{deeplabv3} semantic segmentation network. The EC-CNN block combines a convolutional layer and some gated feature-wise transform(GFT) layer~\cite{featurewise} to utilize the output of EdgeNet to guide the segmentation of the input image.

Wang et al.~\cite{cgan} proposed a thermal infrared pedestrian segmentation algorithm including a conditional generative adversarial network(IPS-cGAN). Nightvision-Net (NvNet)~\cite{nvnet} adopts the network architecture of FCN-8s~\cite{fcn8s} and introduces data refinement (DR), data normalization (DN), and a weighted loss function to improve the segmentation performance.

Multi-level correction network (MCNet)~\cite{mcnet} proposes a multi-level attention module (MAM) to solve the problem of low resolution and blurred edges. Feature Transverse Network (FTNet)~\cite{ftnet} is an end-to-end trainable convolutional neural network architecture that employs an encoder-decoder structure and an edge-guided component for reliable pixel-level classification. It utilizes residual units~\cite{resnet} and a ResNeXt~\cite{resnext} based encoder network to provide thermal imaging image features at different resolutions.

\section{METHOD}

\subsection{Computational Principle}
According to Planck's law, all natural and man-made objects emit infrared radiation at a given temperature (T), with the intensity depending on the object's emissivity. The heat Radiance equation is given in Eq.\ref{Eq1}.

\begin{equation}
\label{Eq1}
S_{\alpha\nu}=e_{\alpha\nu}B_{\nu}(T_{\alpha})+[1-e_{\alpha\nu}]X_{\alpha\nu},
\end{equation}
Where $S_{\alpha\nu}$ represents the infrared radiant heat signal of the object $\alpha$ captured by the sensor at wave number $\nu$, and e represents the emissivity of the object, which is highly correlated with the material of the object. $T_{\alpha}$ represents the temperature of the object. $B_{\nu}$ is Planck's formula for blackbody Radiance, see Eq.\ref{Eq2}. $X_{\alpha\nu}=\sum_{\beta\neq\alpha}V_{\alpha\beta}S_{\beta\nu}$ The $V_{\alpha\beta}$  is the thermal illumination factor, which reflects what percentage of the heat signal reflected onto an object can be collected by the sensor. Since there are many objects in the environment, it is reflected in the equation as cumulative, representing the total amount of heat signal reflected off the objects in the environment.

\begin{equation}
\label{Eq2}
B(\nu,T)=\frac{2h\nu^3}{c^2}\frac1{\frac{h\nu}{ek_BT}-1}
\end{equation}

Notice that the infrared radiant heat signal $S_{\alpha\nu}$ in the equation can be expressed in terms of three physical quantities, i.e. the temperature T, the emissivity e, and the texture X. Therefore, we decompose these physical quantities from the infrared radiant heat signal and display them. There is a problem in this process, which is referred to in the HADAR~\cite{hadar} as TeX degeneracy, in that completely different temperatures (T), emissivities (e), and textures (X), which combine to produce the same thermal Radiance signal S, may be the same. That is, the behavior of solving for T, e and X separately by means of the thermal Radiance signal S is highly pathological inverse problem solving, and there may be an infinite number of solutions.

Solving highly pathological inverse problems often requires a priori knowledge. In real-world scenarios, most objects, except for a few natural ones, are man-made with consistent industrial standards. That is, their emissivity is highly consistent if they belong to the same material. The material of the object in the dataset is known, and we find the surface emissivity e(m) of the material m in the corresponding longwave infrared(LWIR) band through the NASA JPL ECOSTRESS spectral library~\cite{nasa}. Therefore, this method introduces the surface emissivity e(m) under the LWIR band of the target scene object as a priori.

Once e(m) is known, it follows from Eq.\ref{Eq1} that only S and e will survive when we take the derivative of the wavelength $\nu$:

\begin{equation}
\label{Eq3}
[S/(1-e)]^{'}=[e/(1-e)]^{'}B_{\nu}(T)
\end{equation}

here, the subscript is ignored in Eq.\ref{Eq3} to prevent confusion, where prime indicates derivative with respect to wavenumber $\nu$. Since S is observable and e is known, the temperature T can be obtained by solving Eq.\ref{Eq2}. Transforming Eq.\ref{Eq1} yields:

\begin{equation}
\label{Eq4}
V_0=\frac{S_\nu-e_\nu B_\nu(T)}{(1-e_\nu)B_\nu(T_0)}
\end{equation}

Once e and T are known, V can be found according to Eq.4. Finally, X is found according to $X=\int V_{0}B_{\nu}(T_{0})\mathrm{d}\nu$. Thus, we obtain the three physical quantities T, e and X.

\subsection{Pseudo-TeX Vision}

In TeX Vision, it is necessary to use the input of hyperspectral thermal cubes to solve the emissivity of materials, which is costly and more demanding to use for the time being. Therefore, we use the Pseudo-TeX Vision approach to introduce the surface emissivity of the material as a priori and extend its application to thermal datasets of scenes without spectral resolution.

After obtaining the temperature T, emissivity e and texture X, they are mapped into the HSV color space. In the commonly used RGB color space, each channel represents a light intensity, while each channel in the HSV color space used in Pseudo-TeX Vision has a physical meaning, and the Hue channel has a strong correlation with the everyday semantic information of an object. For example, blue color usually indicates water or sky, green color usually indicates grass or leaves, and red color represents heat generating objects. In order to reconstruct or simulate visible RGB images using thermal Radiance signals, in Pseudo-TeX Vision, the material class m of the object is displayed in the Hue channel, which can also be viewed as e since they are highly correlated, the temperature T of the object is displayed in the Saturation channel, and the texture X of the image is displayed in the Value channel.

We perform a pixel-wise operation. First, the infrared image is labeled using labelme, and the labeled data is used to fine-tune the infrared image semantic segmentation network FTnet~\cite{ftnet}, then inference is performed to obtain the semantically segmented image. According to the labels of the semantic segmentation results, each pixel is labeled with its material class, its corresponding emissivity is found according to the NASA JPL ECOSTRESS spectral library~\cite{nasa}, and the temperature T and texture X are computed according to the above equations, and then mapped into HSV space according to the corresponding relationship, which facilitates the subsequent synthesis of novel view using Neural Radiance Field.
 
\subsection{Neural Rendering}

The core principle of Neural Radiance Field is to model the radiation field of a 3D scene by using MLP to generate realistic novel view images. It uses volume rendering to synthesize volume density and color in 3D space combined into ray tracing techniques.

In the process of predicting volume density and color, for each sampled point on the ray, NeRF's MLP accepts the 3D spatial coordinates x and the viewing direction d of the point, and outputs the volume density   and color  of that point through the neural network:

\begin{equation}
\label{Eq5}
(\sigma,\mathbf{c})=\mathrm{MLP}(\mathbf{x},\mathbf{d})
\end{equation}

The $\sigma$ denotes the density of the point and $\mathbf{c}$ denotes the color value of the point. For each ray, the final color $c(r)$ is computed by accumulating multiple sample points. For ray $r(t)$, the final color is calculated by the following volume rendering equation:

\begin{equation}
\label{Eq6}
\mathbf{C}(\mathbf{r})=\sum_{i=1}^NT_i(1-\exp(-\sigma_i\delta_i))\mathbf{c}_i
\end{equation}
Where $T_{i}=\exp\biggl(-\sum_{j=1}^{i-1}\sigma_{j}\delta_{j}\biggr)$ denotes the transmittance of the light from the first point to the $ith$ point, describing the probability that the light is not blocked by the previous point. $\sigma_{i}$ is the volume density of the $ith$ sampling point. $\mathbf{c}_i$ is the color of the $ith$ sample point. $\delta_i$ is the distance between neighboring sampling points.

In the original NeRF, the output of the MLP consists of the volume density $\sigma$ and the color $c=(r,g,b)$. Since Pseudo-TeX vision maps infrared images to HSV color space, by changing the RGB representation to HSV representation, we expect the color branch output of the MLP to be (H, S, V), with the volume density remaining unchanged:

\begin{equation}
\label{Eq7}
(\sigma,H,S,V)=MLP(x,d)
\end{equation}
where $\sigma$ remains the volume density, H stands for Hue, which indicates the type of color, usually in the [0,1] interval, S stands for Saturation, which indicates the intensity of the color, also in the [0,1] interval, and V stands for Value, which indicates the brightness of the color, usually in the [0,1] interval. During volume rendering, $c_{i}=(H_{i},S_{i},V_{i})$ , and the rendering formula calculates the color contribution of each point from the HSV value:

\begin{equation}
\label{Eq8}
C_{HSV}(r)=\sum_{i=1}^NT_i(1-exp(-\sigma_i\delta_i))(H_i,S_i,V_i)
\end{equation}
where $(H_{i},S_{i},V_{i})$ are the hue, saturation and values of the point.

\begin{figure}[!t]
    \centering

    \includegraphics[width=.95\linewidth]{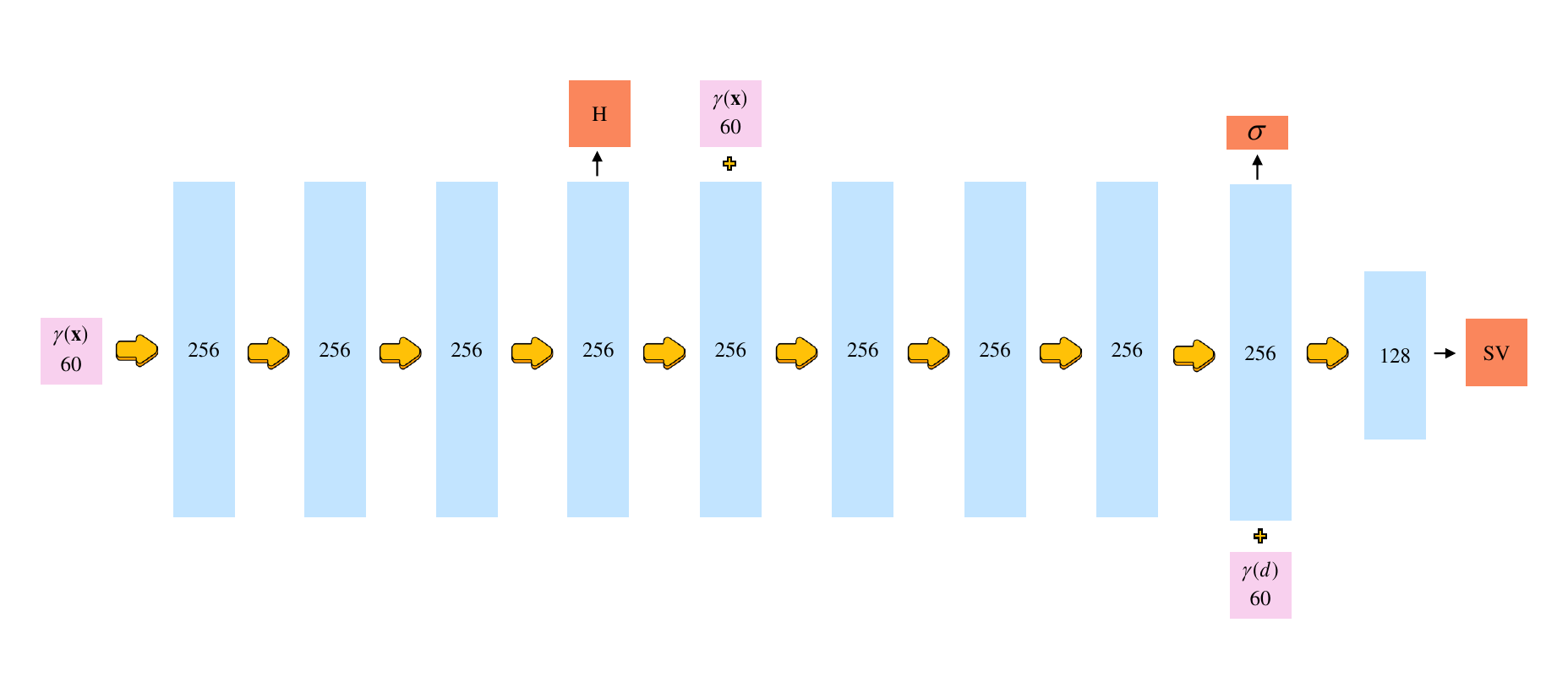}
    \vspace{-3mm}
    \caption{Diagram of TeX-NeRF network architecture.}
    \label{fig3}
\vspace{-4mm}
\end{figure}

In addition, hue represents the type of color, and its changes are usually periodic and smooth, so we use a relatively small network complexity to capture this relatively simple change. As shown in Fig.\ref{fig3}, we output Hue in the shallow network. Since the values of H, S, and V are in the range of [0,1], in order to ensure that the output of the network is within this interval, a sigmoid activation function is used in the output layer. The sigmoid function maps the input to the [0,1] interval and ensures that the H, S, and V values are reasonable.

When the output space is changed to HSV, the loss function needs to be improved accordingly. Considering that hue is a cyclic space, it usually takes values between [0,1]. Therefore, the loss of color phase needs to consider its periodicity, and directly calculating the difference between the two may ignore the case of spanning 0 and 1. For this problem, the following loss function is used:

\begin{equation}
\label{Eq9}
\mathcal{L}_{H}=\min\left(\mid\mathbf{H}^\text{pred}-\mathbf{H}^\text{gt}\mid,1-\mid\mathbf{H}^\text{pred}-\mathbf{H}^\text{gt}\mid\right)
\end{equation}

This loss will ensure that the periodicity of the hue is captured correctly. For Saturation and Value, whose components do not suffer from the above problem, the loss can be calculated using the standard MSE. So, the final loss function can be written as:

\begin{equation}
\label{Eq10}
\mathcal{L}=\frac{1}{N}\sum_{i=1}^{N}\left(\mathcal{L}_{H}+\parallel\mathbf{S}_{i}^{\mathbf{pred}}-\mathbf{S}_{i}^{\mathbf{gt}}\parallel^{2}+\parallel\mathbf{V}_{i}^{\mathbf{pred}}-\mathbf{V}_{i}^{\mathbf{gt}}\parallel^{2}\right)
\end{equation}

\section{EXPERIMENTS}

In this section, we first introduce the datasets used for the experiments, and based on it, we do a comprehensive evaluation and validation, including the quality of the novel view synthesis, the quality of the 3D scene reconstruction, and the evaluation of the accuracy of the synthesized novel view temperature. The rationality and validity of our method is demonstrated by comparing it with a variety of excellent novel view synthesis methods.

\subsection{Datasets}
\noindent{\bf 3D-TeX Dataset.}
Due to the lack of 3D reconstruction datasets for infrared images, the 3D-TeX dataset was acquired for this experiment. 3D-TeX consists of a set of high quality infrared images and the corresponding Pseudo-TeX vision data. An infrared camera model iRAY LGCS121 was used for data acquisition, which has a resolution of 1280 x 1024 pixels, an image element spacing of 12 µm, a detector frame rate of 30 Hz, a response band of 8 to 14 µm, and a NETD (Noise Equivalent Temperature Difference) of less than or equal to 40 mK. The dataset includes three different objects: a sculpture made of plaster, a plastic container half-filled with hot water, and a heating table that maintains a constant temperature when powered on.  The dataset covers a wide range of temperatures and provides detailed information on the emissivity of the materials, ensuring that the data is comprehensive and reliable. This information is detailed in a dataset that will be made publicly available for in-depth analysis and application by interested researchers.

\noindent{\bf ThermalMix Datasets.} 
ThermalMix\cite{thermalnerf} is a multi-view, object-centered dataset that contains well-aligned RGB and thermal imaging images captured from six common objects. Thermal infrared images of three of these scenes, hands, pans, and laptops, are selected for this experiment to illustrate the validation of the method. Among these three datasets, the hand dataset contains 42 infrared images with a temperature range of 22.1 to 33.6°C, the pan dataset contains 82 infrared images with a temperature range of 19.4 to 129.8°C, and the laptop computer dataset contains 93 infrared images with a temperature range of 21.6 to 40.2°C.

 \begin{table*}[h]
\caption{Evaluation of novel view synthesis quality and temperature estimation accuracy in different scenarios.}
\vspace{-3mm}
\label{table1}
\begin{center}
\begin{tabular}{cccccccc}
\hline
Metric                 & Method                  & Scene 1       & Scene 2       & hand          & laptop        & pan           & Avg           \\ \hline
\multirow{5}{*}{PSNR$\raisebox{0.3ex}{$\uparrow$}$}
  & instant-ngp             & 21.01         & 21.69         & 16.99         & 14.00         & 18.21         & 18.38         \\
                       & mip-nerf                & 20.74         & 19.88         & 16.31         & 17.62         & 19.59         & 18.83         \\
                       & nerfacto                & 18.39         & 19.43         & 14.42         & 16.46         & 20.18         & 17.78         \\
                       & splatfacto              & 21.90         & 24.05         & 18.36         & \textbf{21}   & 23.87         & 21.84         \\
 & \textbf{TeX-NeRF(ours)} & \textbf{24.97} & \textbf{32.72} & \textbf{20.01} & 20.91 & \textbf{27.21} & \textbf{25.16} \\ \hline
\multirow{5}{*}{SSIM$\raisebox{0.3ex}{$\uparrow$}$}  & instant-ngp             & 0.70          & 0.55          & 0.58          & 0.48          & 0.64          & 0.59          \\
                       & mip-nerf                & 0.66          & 0.54          & 0.60          & 0.53          & 0.67          & 0.60          \\
                       & nerfacto                & 0.65          & 0.54          & 0.51          & 0.52          & 0.62          & 0.57          \\
                       & splatfacto              & 0.78          & 0.53          & 0.55          & \textbf{0.65} & \textbf{0.84} & 0.67          \\
                       & \textbf{TeX-NeRF(ours)} & \textbf{0.83} & \textbf{0.88} & \textbf{0.65} & 0.62          & 0.72          & \textbf{0.74} \\ \hline
\multirow{5}{*}{LPIPS$\raisebox{0.3ex}{$\downarrow$}$} & instant-ngp             & 0.69          & 0.81          & 0.62          & 0.81          & 0.44          & 0.67          \\
                       & mip-nerf                & 0.67          & 0.61          & 0.43          & 0.56          & 0.41          & 0.54          \\
                       & nerfacto                & 0.66          & 0.59          & 0.46          & 0.52          & 0.42          & 0.53          \\
                       & splatfacto              & 0.51          & 0.54          & 0.30          & \textbf{0.37} & \textbf{0.27} & 0.40          \\
                       & \textbf{TeX-NeRF(ours)} & \textbf{0.33} & \textbf{0.24} & \textbf{0.23} & 0.39          & 0.32          & \textbf{0.30} \\ \hline
\multirow{5}{*}{MAE$\raisebox{0.3ex}{$\downarrow$}$}   & instant-ngp             & -             & -             & 3.75          & 4.46          & 2.37          & 3.53          \\
                       & mip-nerf                & -             & -             & 4.52          & 3.30          & 1.75          & 3.19          \\
                       & nerfacto                & -             & -             & 4.33          & 3.74          & 1.89          & 3.32          \\
                       & splatfacto              & -             & -             & 2.28          & 1.81          & 1.75          & 1.88          \\
                       & \textbf{TeX-NeRF(ours)} & -             & -             & \textbf{1.90} & \textbf{1.63} & \textbf{0.98} & \textbf{1.50} \\ \hline
\end{tabular}
\vspace{-6mm}
\end{center}
\end{table*}

\subsection{Experimental setup and Evaluation metrics}
When using infrared images for 3D reconstruction, the ghosting effect results in missing image texture, which poses a challenge for NeRF-based 3D reconstruction. Since NeRF requires accurate camera pose information, pose reconstruction algorithms such as COLMAP~\cite{colmap} can often only restore about one-third to one-half of the camera pose when processing infrared images, which in turn affects the reconstruction quality and even leads to reconstruction failure. In order to solve this problem, the Pseudo-TeX vision method is used to preprocess the infrared images. This preprocessing step can effectively enhance the texture information in the image, so that COLMAP can successfully recover all the camera poses, thus providing more reliable input data for the subsequent 3D reconstruction.

In order to fully illustrate the effectiveness of the method proposed in this paper, we perform novel Pseudo-TeX view synthesis on 3D-TeX and ThermalMix datasets and perform quality verification on the synthesized novel views. The temperature accuracy of the novel views is evaluated on ThermalMix. In the experiments, based on the high-quality open-source project nerfstudio~\cite{nerfstudio}, the TeX-NeRF method proposed in this paper is compared in full detail with methods with good novel view synthesis as well as 3D reconstruction results, including instant-ngp~\cite{instantngp}, mip-nerf~\cite{mipnerf}, nerfacto, splatfacto.

In our experiments, we used a variety of evaluation metrics to measure the quality and temperature accuracy of the novel view synthesis to ensure the objectivity and reliability of the results. First, in measuring the visual quality of novel view synthesis, we use three common image quality evaluation metrics: PSNR, SSIM, and LPIPS. PSNR is a pixel-level-based metric for evaluating the difference between a synthesized image and a reference image, which reflects the reconstruction accuracy of the image, with higher values indicating better synthesis quality. However, PSNR only considers the overall pixel value difference of the image, which cannot fully reflect the image quality perceived by the human eye. To compensate for the shortcomings of PSNR, we introduce SSIM, which evaluates the similarity between a synthesized image and a reference image by comparing the brightness, contrast and structural information of the images. Unlike PSNR, SSIM is more in line with the perception of the human visual system and can better capture the changes in local details in the image. In addition, to further evaluate the differences in the perceptual level of the synthesized images, we use the LPIPS metric. LPIPS is based on VGG16~\cite{vgg} and quantifies the perceptual quality differences by extracting high-level features of the images. It is an effective measure of the difference in image quality as perceived by the human eye, and LPIPS is an important complementary metric especially when dealing with images with complex textures and details.

In addition to the visual quality, we also evaluated the temperature accuracy of the novel view synthesis. To evaluate the temperature accuracy in the synthesized viewpoint, we used MAE as the main evaluation metric.MAE directly reflects the accuracy of temperature prediction by calculating the temperature difference between each pixel in the synthesized image and the reference image.

\begin{figure*}[t]
    \centering
    \includegraphics[width=.85\linewidth]{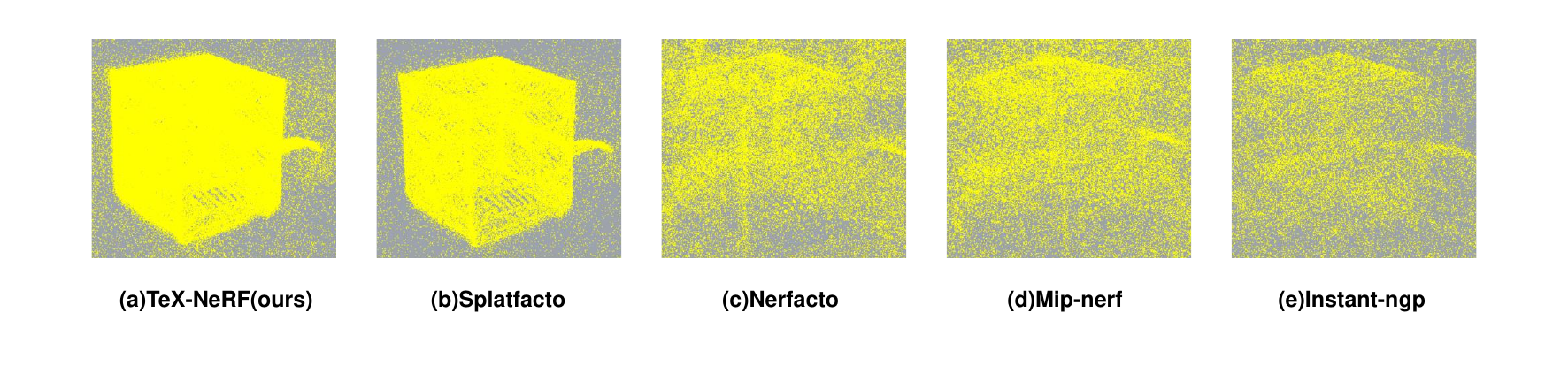}
    \vspace{-6mm}
    \caption{Comparison of 3D scene reconstruction point clouds using TeX-NeRF and other methods on constant temperature heating table images processed by Pseudo-TeX Vision.}
    \label{fig4}
\vspace{-4mm}
\end{figure*}

\subsection{Infrared scenes novel view synthesis}

In the experiment, we used the above-mentioned evaluation metrics to comprehensively evaluate the training effects of the TeX-NeRF method and the comparative method on the Pseudo-TeX vision image sequences generated by the two datasets. The temperature accuracy of synthesizing novel views was evaluated on the ThermalMix dataset, which provides accurate temperature values.

The experiments were first conducted on our own 3D-TeX dataset, which includes scene 1 (sculpture and container) and scene 2 (constant temperature heating table). In addition, we validated the method on the hand, laptop, and pan scenes in the public dataset ThermalMix. We used various methods to validate on the dataset in turn and calculated the average of the evaluation indicators of each method in these scenes, as shown in TABLE \ref{table1}. It can be seen that the TeX-NeRF method not only has a great advantage in the quality of novel view synthesis, but also significantly outperforms other methods in temperature estimation accuracy.

\subsection{3D scene reconstruction}

In this study, the quality of 3D scene reconstruction is evaluated by point cloud density and completeness. Point cloud density directly reflects the ability to capture geometric details when reconstructing a scene. By comparing the point cloud density of different areas, especially those with complex structures or less textures, the performance of the reconstruction methods in different scenes can be judged. Completeness is another key quality evaluation indicator, which refers to whether the point cloud model fully covers all areas in the scene. By visually evaluating the completeness of the point cloud model, it can be found whether there are geometric omissions or holes in the scene. A complete point cloud model should be able to completely and accurately reconstruct all objects and surface details in the scene without losing key parts. Compared with other methods, our method has significant advantages in scene coverage. It can fully reproduce the details of the infrared scene while ensuring geometric accuracy. Fig.\ref{fig4} shows the comparison between our method and several other common methods in terms of the quality of reconstructed point clouds of 3D infrared scenes on a constant temperature heating table in the 3D-TeX Dataset. It can be clearly seen from the figure that our method performs better in point cloud density and completeness, and the generated point cloud is not only more detailed, but also has higher coverage. This comparison further verifies the effectiveness of our method in 3D reconstruction tasks, especially in infrared scene reconstruction.

\section{CONCLUSION}

In this paper, we propose TeX-NeRF, a 3D reconstruction method based on infrared images. We utilize Pseudo-TeX Vision to preprocess infrared images, extracting temperature (T), emissivity (e), and texture (X) information from thermal radiation data and mapping them to the HSV color space. This approach overcomes the ghosting effect in infrared images and restores detailed texture information. The processed images are then fed into TeX-NeRF for training, yielding high-quality 3D infrared scene reconstructions. Experimental results demonstrate that our method surpasses existing techniques in novel view synthesis quality and temperature estimation accuracy. In the future, we aim to enhance the computational efficiency of TeX-NeRF and explore its applications in areas such as robotic night vision and autonomous driving.


\bibliographystyle{IEEEtran}
\bibliography{IEEEabrv,bibfile}

\end{document}